\documentclass[10pt, conference]{IEEEtran}
\IEEEoverridecommandlockouts
\usepackage{cite}
\usepackage{amsmath,amssymb,amsfonts}
\usepackage{algorithmic}
\usepackage{graphicx}
\usepackage{textcomp}
\usepackage{flushend}
\usepackage{hyperref}
\usepackage[letterpaper, top=0.75in, bottom=1.1in,left=0.65in, right=0.68in]{geometry}

\begin{document}

\title{IoT Security*\\
{\footnotesize \textsuperscript{*}Note: Sub-titles are not captured in Xplore and
should not be used}
\thanks{Identify applicable funding agency here. If none, delete this.}
}

\title{Calibrated RF-Fingerprinting Under Interference With Heterogeneous Transmission Protocols}

\author{\IEEEauthorblockN{Tariq~Abdul-Quddoos,~Xiangfang Li, Lijun Qian}
\IEEEauthorblockA{CREDIT Center and Department of Electrical and Computer Engineering  \\
Prairie View A\&M University, Texas A\&M University System  \\
Prairie View, TX 77446, USA \\
Email: \{tabdulquddoos, xili, liqian\}@pvamu.edu}
}


\maketitle

\begin{abstract}
Radio Frequency(RF)-Fingerprinting is a spectrum monitoring technique that identifies specific transmitters based on hardware impairments imprinted within the emitted signal. Although widely researched, studies almost exclusively consider scenarios where only one transmitter is emitting at a time, limiting real world applicability. In this work, we further the study of RF-Fingerprinting by considering co-channel interference, with multiple emitted signals interfering with each other, overlapping in time and frequency. Specifically, we formulate this problem as a multi-label classification problem and  employ a 1D convolutional neural network (CNN). Furthermore, the models are calibrated such that the confidence thresholds for the label probabilities are derived, with guarantees on the upper bound on the average number of False Negatives, providing a degree of confidence in not missing a true spectrum policy violation. The proposed method is validated using real world data from the POWDER 5G testbed on devices transmitting 802.11a(Wi-Fi), 4G LTE, and 5G NR waveforms. The results show accuracy as high as $97\%$ and as low as $73\%$ after calibration depending on channel conditions. Also calibrating for various average false negatives upper bounds achieves micro recall scores of approximately (1 - calibrated false negatives)  with the calibration robust to out-of-distribution interference, demonstrating the potential of the proposed method in a realistic high contention wireless environment. 
\end{abstract}


\section{Introduction}
Radio frequency (RF) fingerprinting refers to methods for identifying transmitter hardware based on a received signal. Hardware imperfections present as features that are embedded in a signal forming a fingerprint that is unique to that device. These imperfections are present within the analog components such as digital-to-analog converters, mixers, and power amplifiers, differentiating radio devices even if their manufacturer/make/model are identical~\cite{NoRadioLeft}. Features can present themselves as imbalance in the In-Phase and Quadrature components of the signal and can also present as distortions in the phase and amplitude of the signal. Deep learning models have shown to be able to extract these features and accurately classify the transmitting hardware given that all signal samples are from the same receiver~\cite{IoTFingerprinting_Qian2018,IoTFingerprinting_Qian2019}.

The majority of the fingerprinting literature focused on the problem of identifying the transmitter in scenarios when only one transmitter is active at a time. However, multiple transmitters may be active, overlapping in both time and frequency and using different protocols, especially with the increasing number of wireless devices and services. Such a scenario would require the fingerprinting algorithm to separately distinguish and extract the signatures of each transmitter from a single received signal~\cite{compsurvey}. The interference due to overlapping transmissions can be broadly categorized as adjacent channel and co-channel interference. The main difference between these two is that the former occurs between adjacent frequency bands while the latter occurs in the same frequency band~\cite{INFSUP}. Standard solutions involving identifying a source or separating overlapping signals involve filtering out interference by masking irrelevant parts of the time-spectrum grid~\cite{SourceSep}, using multi-antenna capabilities to focus on specific spatial directions~\cite{multipleEmitter}, and statistical methods such as blind source separation~\cite{BlindSource}. 

\begin{figure}[]
	 \centering
    	 \includegraphics[width=0.45\textwidth]{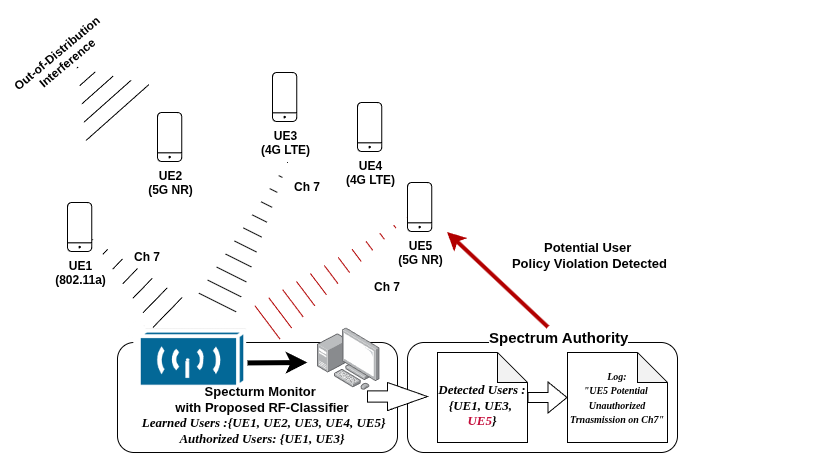}
      	\caption{Proposed Framework for Unauthorized Transmission in Co-Channel Wireless Environment Classifier}
    \label{fig:Motivation}
\end{figure}

One of the challenges involved in studying the co-channel scenario using data-driven modeling is the availability of a real-world dataset that incorporates multiple active emitters~\cite{compsurvey}. In the propagation of real wireless signal transmissions, each individual component of the overlapping signals experiences diverse propagation loss, which means the scaling among each component signal exists differences~\cite{JCCM}. When applying real-world data to data-driven modeling there exist several sources of uncertainty broadly categorized into aleatoric and epistemic uncertainty. Aleatoric uncertainty arises from unaccounted factors, such as channel conditions or receiver hardware impairments and is also known as an irreducible error that cannot be remediated with more data~\cite{quantifyinguncertainty}. Epistemic uncertainty arises from the models lack of knowledge and can be remedied with more data and statistical tools~\cite{quantifyinguncertainty}. \emph{Quantifying uncertainty is important for having a degree of confidence of the fingerprinting models performance when deployed especially for identifying policy violations in high contention wireless environments, see Fig.\ref{fig:Motivation}}. In this work, the technique of Conformal Prediction~\cite{CP} is applied as a powerful uncertainty calibration technique due to it being model agnostic and distribution-free. Specifically, we propose the use of Conformal Risk Control~\cite{CRC}, such that performance guarantees can be configured and controlled regarding logical risk a model may be prone to in its predictions.

The contributions of this study include
\begin{itemize}
    \item  A real-world dataset for RF-Fingerprinting in a Co-Channel scenario created with heterogeneous transmission protocols and collected from the POWDER testbed; 
    \item The RF fingerprinting problem is formulated as a multi-label classification problem and a lightweight multi-label RF classifier based on a 1D CNN is proposed that is applicable at the network edge while achieving high accuracy; 
    \item An uncertainty-aware post-processing mechanism is proposed using conformal risk control for controlling false negative risk. This scheme provided much needed  confidence of the fingerprinting model performance for identifying potential policy violations by known users in a shared spectrum.
\end{itemize}

The remainder of the paper is structured as follows. Section II reviews related works. Section III describes the dataset collection process and problem formulation. The proposed method is explained in Section IV. Experimental results and analysis are given in Section V. Section VI concludes the study.

\section{Related Work}
While there is a large body of literature addressing Radio Frequency(RF)-Fingerprinting, there is few that specifically address the problem with multiple transmitters interfering with each other. One recent work and closest to our approach is~\cite{specificemitter}, where they use a Convolutional Neural Network to extract signal features and map them to multiple transmitters. They use a simulation to generate data, considering different SNR levels and various amounts of overlap in the operating bands of transmitters. An earlier work addressing co-channel interference among emitters is~\cite{RFFemto}, and uses frequency domain analysis combined with traditional discriminatory classifiers such as K-Nearest Neighbor(KNN) to perform transmitter identification. This work specifically addresses the problem of overlapping signals in femtocells with data collected using a lab setup mimicking signal storms caused by idle cell signals. The work in~\cite{JCCM} also does classification with overlapping signals but classify the modulation type instead of the transmitter. 

A set of literature close to the co-channel scenario with overlapping transmitters is the scenario where there is ambient signals co-channel with a single signal of interest, common with technologies like bluetooth and Wi-Fi. In~\cite{RFUAV} they address the problem of classifying Unmanned Aerial Vehicles(UAVs) based on the RF-Fingerprints in the presence of ambient co-channel interference from Wi-Fi and bluetooth sources. This work used a multi-stage detector to first detect if a signal was from a UAV related source or noise due to co-channel interference, then classifying which UAV source it came from. Their experimental setup consists of real-world RF signals from UAV controllers, Wi-Fi devices, and bluetooth devices. In~\cite{MutliTaskRF_Atten} they consider multi-task classification under out-of-distribution interference, classifying the transmitting device and the protocol between bluetooth and Wi-Fi. They also gather data using an indoor lab setup, with data collected over a period of months capturing distribution shift. 


To the best of our knowledge, previous research utilizing deep neural networks in RF fingerprinting has predominantly focused on classifying single transmitter, or if classifying multiple transmitters overlapping in time and frequency, data often comes from simulation with few that use real-world data. In contrast, our research employs a lightweight 1D CNN to achieve high classification accuracy when multiple transmitters are interfering with each other and with heterogeneous transmission protocols in an over-the-air lab setting. In addition, an uncertainty-aware post-processing mechanism is proposed using conformal risk control to provide the much needed confidence of the fingerprinting model performance for identifying active devices and potential policy violations for known spectrum users.

\section{Dataset Collection and Problem Formulation}
\subsection{Dataset Collection}
\label{sec:dataset}

\begin{figure}
	 \centering
    	 \includegraphics[height=0.12\textheight]{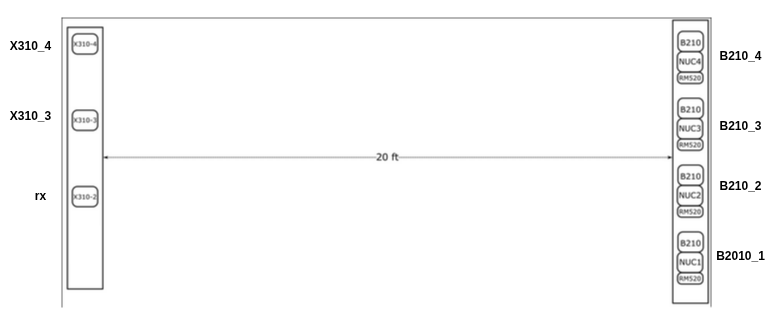}
      	\caption{POWDER Indoor Over-the-air lab setup. Adapted from ~\cite{POWDERCite}.}
    \label{fig:POWDEROtaLab}
\end{figure}

The data used in this work is a real-world dataset collected using the indoor over-the-air lab at the University of Utah's POWDER testbed~\cite{POWDERCite}. The experimental setup shown in Fig.\ref{fig:POWDEROtaLab} consist of seven software defined radios(SDR) with one being  used as the receiver and six being used as transmitters. The receiving radio as well as two transmitting radios are NI/Ettus X310 USRP's and the remaining four are NI/Ettus B210 USRP's. The X310 has a receiver gain range of 0 - 31.5 with the gain set to 25 across all experiments. The transmitting gain range on the X310 is 0 - 31.5 and for the B210 the range is 0 - 89, with gains set to 60\%, 70\%, 80\%, and 90\% of the max range. For the X310 the exact gains are  28.4, 25.2, 22.1, 18.9 and for the B210 they are 80.8, 71.8, 62.9, 53.9. Transmitted waveforms are generated by the MATLAB wireless waveform generator and consist of 5G NR, 4G LTE, and 802.11a(Wi-Fi), all with a bandwidth of 20MHz.

The data is collected in three rounds with the corresponding protocol assignment for each radio shown in Table \ref{tab:protocol_assignment}.
\begin{table}[h!]
\fontsize{7}{8}\selectfont
\caption{Transmission Protocol Assignment}
\label{tab:protocol_assignment}
\begin{tabular}{|c|c|c|c|}
\hline
\textbf{Round} & \textbf{X310\_3 \& X310\_4} & \textbf{B210\_1 \& B210\_2} & \textbf{B210\_3 \& B210\_4} \\
\hline
Round 1 & 5G NR & 4G LTE & 802.11a (Wi-Fi) \\
\hline
Round 2 & 4G LTE & 802.11a (Wi-Fi) & 5G NR \\
\hline
Round 3 & 802.11a (Wi-Fi) & 5G NR & 4G LTE \\
\hline
\end{tabular}
\end{table}

For each round four sets of data are collected at each gain percent, where a single set in the round defined in eq.(\ref{eq:dataset_rounds}), consist of captured signals for each permutation of transmitter on/off states. 

\begin{equation}
\mathcal{D}_{r,s} = \{ (x_i(t), \pi_i(\mathcal{Y})) \mid i = 0, \dots, 63, \mathcal{Y}\in\{0,1\}^6\}
\label{eq:dataset_rounds}
\end{equation}
Where r is the round, s is the set, $x(t)$ is the received I/Q samples, $\mathcal{Y} = [y_1,...,y_6]$ is a vector of transmitter states $y_l\in\{0, 1\}$, and $\pi(\mathcal{Y})$ is a permutation of the transmitter states. The total dataset is referred to as the POWDER Co-Channel Protocol(PCP) Dataset and is the union of all the sets together $\mathcal{D}_{PCP} = \bigcup_{r=1}^{3}\bigcup_{s=1}^{4} \mathcal{D}_{r,s}$.

    The carrier frequency used is 2.425 GHz, the sampling rate is 33.33 MS/s, and 20 million samples are collected for every signal. In total the dataset consist of 768 signals where each round consist of 256 signals and each set of experiments in the round captures 64 signals.The dataset can be found at: \textit{https://huggingface.co/datasets/T-Arshad/POWDER\_CoChannel\_Protocol\_Dataset}

\subsection{Signal Model}
\label{sec:signal_model}
The received signal $x(t)$ is represented as the superposition of the active transmitters signals distorted by their transmitting hardware and propagating through their respective channels as shown in eq.(\ref{eq:input_signal}). 
\begin{equation}
x(t)  = \sum_{l \in \mathcal{Y}_{\text{active}}} d_{l}(s_{l}(t))*c_{l}(t) +  w(t)
\label{eq:input_signal}
\end{equation}

Where $\mathcal{Y}_{\text{active}} \subseteq \mathcal{Y}$ is the active transmitter subset, $l$ is the specific label from the active subset, $c_{l}(t)$ represents the channel impulse response, $s_l(t)$ is the I/Q samples, $d_{l}(s_{l}(t))$ represents the distortion function on a transmitted signal, and $w(t)$ is gaussian noise.  The distortion function $d_{l}(s_{l}(t))$ is defined by I/Q imbalance, carrier frequency offset, carrier leakage, spurious tones, and power amplifier(PA) nonlinearity.  

Following~\cite{RFImpairbook}, the effects of I/Q imbalance is defined in eq.(\ref{eq:distortion_func_1}).
\begin{equation}
d'(s(t)) =  \alpha_{tx}(s(t)) + \beta_{tx}(s^*(t))
\label{eq:distortion_func_1}
\end{equation}
\begin{equation}
\alpha_{tx} = \frac{1}{2}[cos(\frac{\Delta\phi}{2}) - j\frac{\Delta{G}}{2}sin(\frac{\Delta\phi}{2})]
\label{eq:distortion_func_11}
\end{equation}
\begin{equation}
\beta_{tx} = \frac{1}{2}[-\frac{\Delta{G}}{2}cos(\frac{\Delta\phi}{2}) + jsin(\frac{\Delta\phi}{2})]
\label{eq:distortion_func_12}
\end{equation}
Where $\Delta{G}$ is the gain imbalance quantifying the difference between the gain in the real component(I) path and quadrature component(Q) path and $\Delta\phi$ is the phase error between the I and Q component paths.
Following~\cite{multipleEmitter} effects of the carrier frequency offset, carrier leakage, and spurious tone are shown in eq.(\ref{eq:distortion_func_2}). 
\begin{equation}
\begin{split}
d''(s(t)) = e^{j 2\pi (f_{tx} - f_{rx})t} \Big( d'(s(t)) + \zeta_{CL} + a_{ST} \, e^{j 2\pi f_{ST} t} \Big)
\end{split}
\label{eq:distortion_func_2}
\end{equation}
The carrier frequency offset is represented as $f_{tx} - f_{rx}$, the carrier leakage is represented by $\zeta_{CL}$ and the spurious tone is represented by $a_{ST}$ and $f_{ST}$.
The power amplifier nonlinearity is represented by the memory polynomial model~\cite{PANonLinear} as shown in eq.(\ref{eq:distortion_func_3}), where $k$ is the order on nonlinearity, $m$ is the memory depth, and $c_{k,m}$ are the model coefficients.

\begin{equation}
d'''(s(t)) = \sum_{k=1}^{K}\sum_{m=1}^{M} c_{k,m}d''(s(t - m))|d''(s(t-m))|^k
\label{eq:distortion_func_3}
\end{equation}


\subsection{Problem Formulation}
\label{sec:Problem}
The goal is to define a data-driven model that takes in received I/Q samples($x(t)$) and returns the states of all learned transmitters($\hat{\mathcal{Y}}$). The problem is formulated as a two stage multi-label classification problem. The first stage consist of a model, $f(x(t); \theta)$ parameterized by learned weights $\theta$, that returns independent probabilities, $\hat{y}_l$, that a particular transmitter $l$ is active in signal $x(t)$ as shown eq.(\ref{eq:cond_classprob}).
\begin{equation}
\hat{y}_{l} = P(y_l = 1 \mid x(t))
\label{eq:cond_classprob}
\end{equation}

The optimal weights $\theta^*$ are found such that conditional marginal distribution between the signal $x(t)$ and the transmitter states $\mathcal{Y}$ is learned. This is done by minimizing the binary-cross entropy loss function across N examples as shown in eq.(\ref{eq:bce}). 
\begin{equation}
\begin{split}
\theta^* = \arg\min_{\theta}\\\left\{ -\frac{1}{N} \sum_{i=1}^{N} \sum_{l\in\mathcal{Y}} \left[ y_{i,l} \log(\hat{y}_{i,l}) + (1 - y_{i,l}) \log(1 - \hat{y}_{i,l}) \right] \right\}
\end{split}
\label{eq:bce}
\end{equation}

In the second stage an uncertainty aware post-processing function $T_{\lambda}(\hat{y})$, is defined that takes in probabilities returned by the model, and returns all active transmitters $\hat{\mathcal{Y}}_{active}$ as defined in eq.(\ref{eq:set_func}). 

  \begin{equation}
         T_{\lambda}(\hat{y}) = \{\ l:\hat{y}_l \geq 1- \lambda \}\
        \label{eq:set_func}
    \end{equation}

The parameter $\lambda$ is derived following the procedure of Conformal Risk Control\cite{CRC}, such that the optimal parameter $\hat{\lambda}$ ensures the expected values of an uncertainty based calibration function $\mathcal{C}(T_{\lambda}(\hat{y}), \mathcal{Y}_{active})$ is upper bounded by a user defined value $\alpha$ as shown in eq.(\ref{eq:CRC_guarantee}).  

    \begin{equation}
  E\!\left[ \mathcal{C}( T_{\hat\lambda}(\hat{y}),\mathcal{Y}_{active} ) \right] \leq \alpha
        \label{eq:CRC_guarantee}
    \end{equation}

The calibration function must be constructed such that it is non-increasing and right continuous, the False Negative Rate(FNR) calibration function shown in equation~(\ref{eq:fnr_loss_func}) satisfies these conditions. With the FNR function we seek to control the rate of false negatives while constructing the smallest possible prediction set.   
    \begin{equation}
        \mathcal{C}_{FNR}\big( T_{\lambda}(\hat{y}),\mathcal{Y}_{active}) = 1 -  \frac{\lvert \mathcal{Y}_{active} \cap T_\lambda(\hat{y}) \rvert}{\lvert \mathcal{Y}_{active} \rvert}
        \label{eq:fnr_loss_func}
    \end{equation}

\section{Proposed Methods}
\label{sec:method}

Details on I/Q samples pre-processing,  model architecture, and empirically defining the post-processing function are given in this section. 
\subsection{Pre-Processing}
For pre-processing the Discrete Fourier Transform(DFT) shown in eq.(\ref{eq:fourier}) is applied to each signal, with a hanning window used and each DFT has a length of 1024.
\begin{equation}
    X_k = \sum_{n=0}^{N-1} x[n] \, e^{-2 \pi j \frac{k n}{N}}, \quad k = 0, 1, \dots, N-1
    \label{eq:fourier}
\end{equation}
Where $x[n]$ is the I/Q sample, N is the total number of samples,   $X_k$ is the frequency component, and  k is the frequency index. The frequency components are then then split into four channels where the channels in this order are the real component, imaginary component, magnitude, and phase.

\subsection{Model Architecture}
A convolutional neural network (CNN) is adopted as the classifier's model architecture. The specific CNN architecture in this work follows the structure of the 1D CNN classifier from~\cite{RFCLASSIFY}. The model takes a 1D sequence with 4 Channels and the sequence is passed through 5 1D convolution layers each with a kernel size of 9, stride of 1 and output channel sizes of 256, 256, 128, 64, and 32. Additionally batch normalization, max pooling with a stride and kernel size of 2, and dropout with a rate of 0.3 are added to each layer. After the convolution layers the feature maps are flattened and passed through 2 feed-forward layers, the first making the flattened sequence into a length of 256 with a dropout rate of 0.3, and the second making the sequence the length of the number of classes. All layers have a ReLu activation function except the last where a Sigmoid function is applied to get independent per-class probabilities. The total number of model parameters is 1.25M taking up only 5 MB of space during inference.   

\subsection{Decision Making Function}
 Following conformal risk control~\cite{CRC} the optimal thresholding parameter $\hat\lambda$ is derived empirically for the function $T_{\hat\lambda}(\hat{y})$. For doing this calibration empirically, given dataset of N calibration points $D_{cal}$  we first define a discrete set of $\lambda$ candidate values, $\Lambda = \{ \lambda_1, ..., \lambda_N\}$, ordered from smallest to largest. For each value in $\Lambda$, the empirical expected value of the calibration function is calculated as defined in eq.(\ref{eq:empirical_expected}),
 \begin{equation}
 \hat{E} = (\mathcal{C}_{FNR_1} + \mathcal{C}_{FNR_2}+...\mathcal{C}_{FNR_N})/N
 \label{eq:empirical_expected}
 \end{equation}
 calculated across all examples in $D_{cal}$. The value of $\hat\lambda$ used in the post processing function is the infimum of the set of all $\lambda\in\Lambda$ satisfying $\hat{E} \leq \alpha$ as defined in equation~(\ref{eq:lamda_select}).   
    \begin{equation}
       \hat{\lambda} = inf\{\lambda \in \Lambda: \hat{E} \leq \alpha \}
        \label{eq:lamda_select}
    \end{equation}

\section{Results}
\label{sec:results}

\subsection{Experimental Setup}
For the experimental setup, signals containing transmissions from 5 radios (X310\_3, X310\_4, B210\_1, B210\_2, and B210\_3) are trained on, with another radio (B210\_4) held out for evaluation acting as a source of out-of-distribution(OOD) interference. From the larger dataset $D_{PCP}$ three smaller datasets, $D_{train}, D_{Cal}, D_{Test}$, are made, where signal examples of length 1024 are sampled from every permutation of the transmitter states. $D_{train}$ consist of 40000 examples without the OOD interference. $D_{Cal}$ consist of sets of 10000 examples for each gain and number of channel occupants without the OOD interference. $D_{Test}$ consist of sets of 10000 examples for each gain and number of occupants, where in each set 5000 do not have the OOD interference and 5000 do have the OOD interference. The model is trained on the entire training set then calibrated and evaluated at each individual gain and number of channel occupants. A learning rate of 0.001, batch size of 256, 25 epochs, and adam optimizer are used for training. Results are shown for the calibration and post-calibration evaluation done at $\alpha$ values 0.05, 0.15, and 0.25. The discrete calibration candidate set $\Lambda$ consists of all $\lambda$ values between 0 and 1 spaced by 0.01. 

\subsection{Calibration Results}
\begin{figure}[h!]
    \centering
\includegraphics[width=0.50\linewidth]{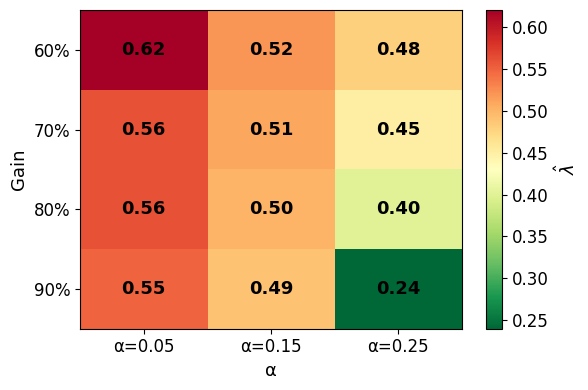} 
    \caption{Gain FNR Calibration Results}
    \label{fig:gain_calib_Result}
\end{figure}

\begin{figure}[h!]
    \centering
\includegraphics[width=0.50\linewidth]{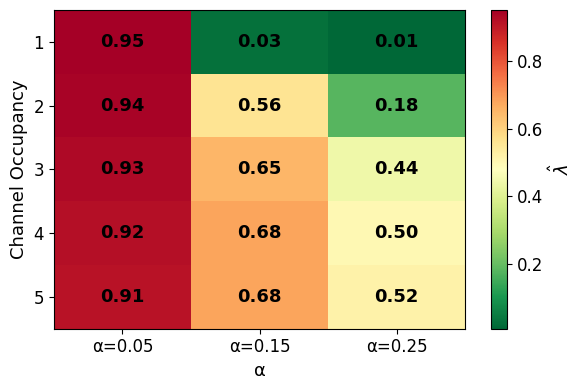} 
    \caption{Channel Occupancy FNR Calibration Results}
    \label{fig:occupancy_calib_Result}
\end{figure}

The results for each derived value of $\hat\lambda$ is shown in Fig.\ref{fig:gain_calib_Result} for each gain and in Fig.\ref{fig:occupancy_calib_Result} for each number of channel occupants. For both calibrations the value of $\hat\lambda$ decreases as $\alpha$ increases. For the gain calibration the value of $\hat\lambda$ decreases as gain increases. For the occupancy calibration the value of $\hat\lambda$ increases as the number of occupants increases when $\alpha$ is 0.15 and 0.25, whereas when $\alpha$ is 0.05, $\hat\lambda$ instead decreases as occupants increase.  

\subsection{Post-Calibration Results}
\begin{figure*}[]
\centering
\includegraphics[width=0.75\textwidth]{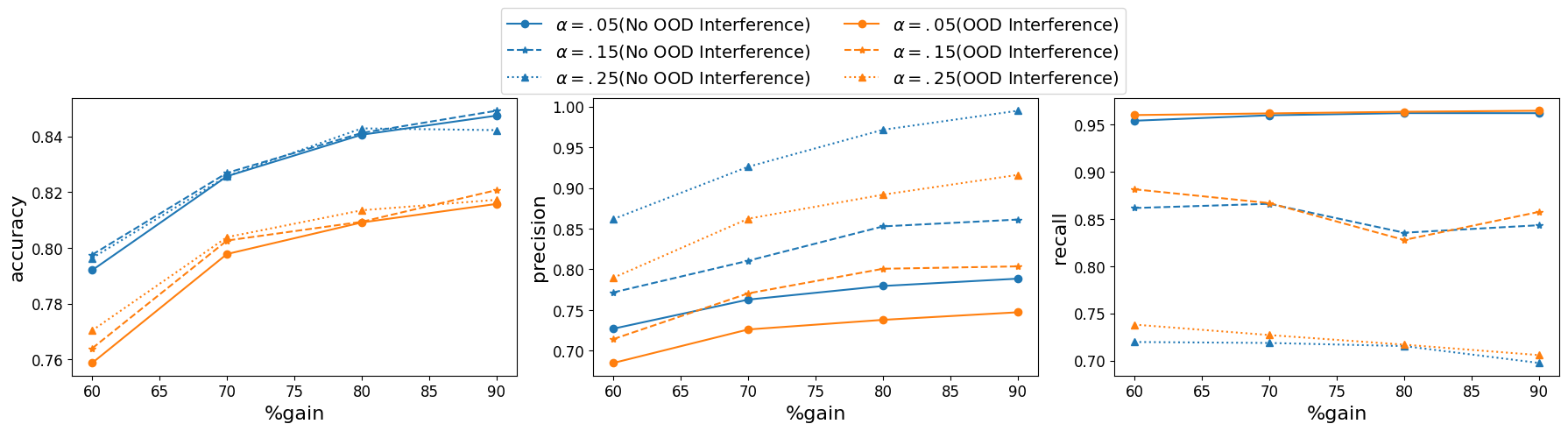}
\caption{Overall Accuracy, Precision, and Recall Across Gain}
\label{fig:gain_overall_eval_Resuls}
\end{figure*}

\begin{figure*}[]
\centering
\includegraphics[width=0.75\textwidth]{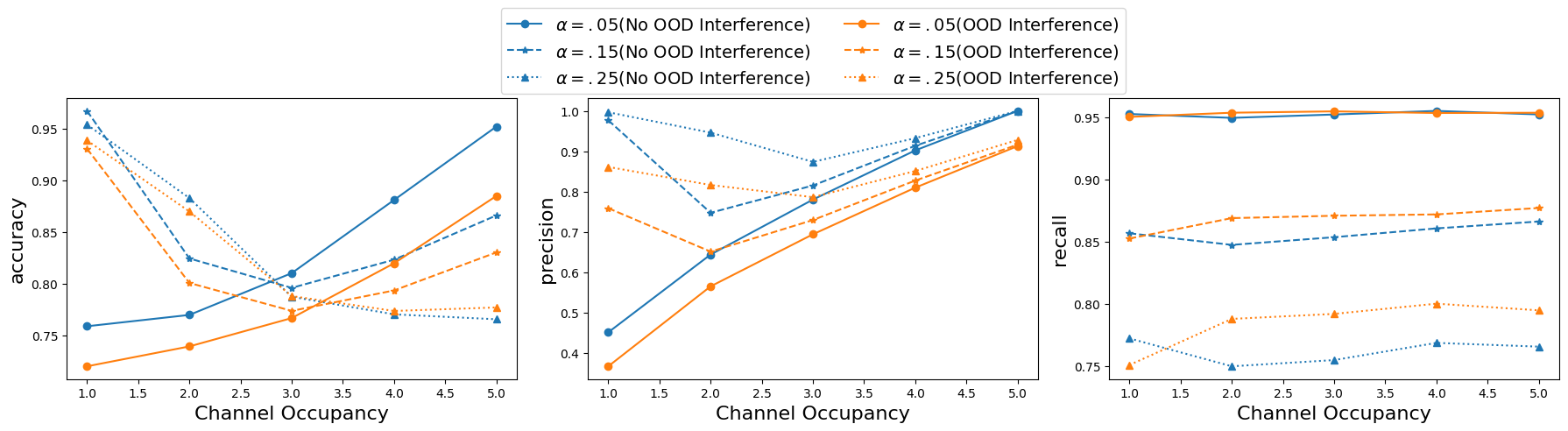}
\caption{Overall Accuracy, Precision, and Recall Across Number of Channel Occupants}
\label{fig:occupancy_overall_eval_Resuls}
\end{figure*}

Post-calibration evaluation for all radios is shown in Fig.\ref{fig:gain_overall_eval_Resuls} with respect to the gain of the device and in Fig.\ref{fig:occupancy_overall_eval_Resuls} with respect to the number of channel occupants, with accuracy, micro-precision, and micro-recall as metrics. The effect of the calibration is shown in the recall scores where each recall is around 1 - $\alpha$. When the OOD interference is introduced the calibration remains consistent with a max loss of performance of 0.04 when the channel occupancy is at 2 and $\alpha$ is 0.25. Across all experiments there is a more significant loss in precision performance when the OOD interference is introduced with a max loss of 0.22 when the channel occupancy is at 1 and $\alpha$ is 0.15. For the accuracy performance is as high as 0.97 and as low as 0.73 depending on channel conditions with a loss of performance as much as 0.07 when the OOD interference is introduced.  


\begin{figure*}[]
\centering
\includegraphics[width=0.95\textwidth]{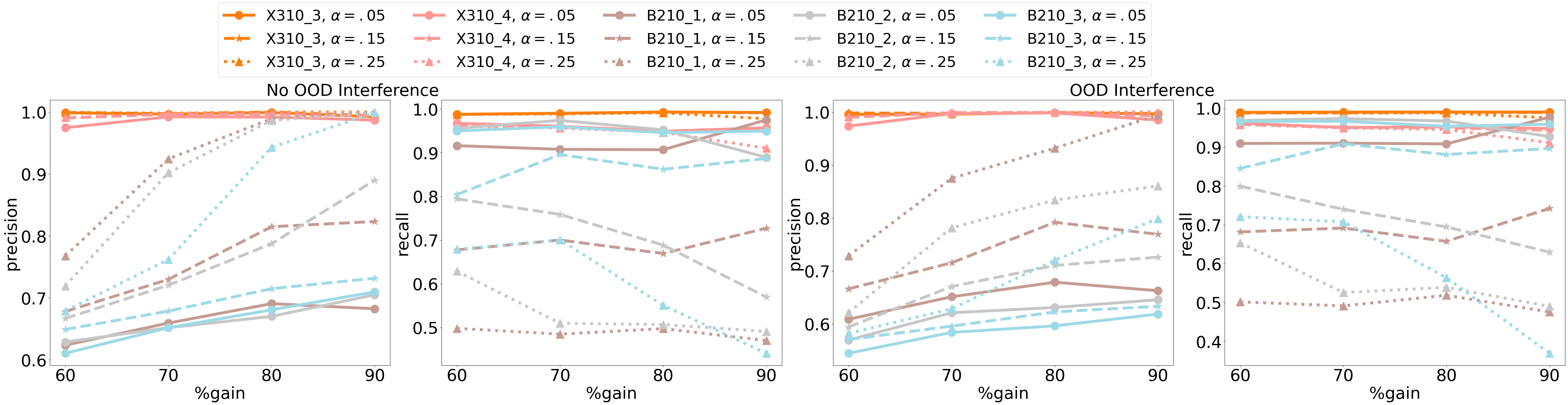}
 	\caption{Individual Radio Precision and Recall}
\label{fig:class_Results}
\end{figure*}

Results for each individual device across gains is shown in Fig.\ref{fig:class_Results} and evaluated using precision and recall. The X310 radios show the best performance regardless of gain, $\alpha$ or the OOD interference, with X310\_3 having a precision and recall around 0.99 across all experiments. For the X310\_4 the precision ranges from 0.97-0.99 and recall ranges from 0.94 - 0.97. For the B210 radios their precision increases as gain and $\alpha$ increases, with a loss in performance of as much as 0.24 when the OOD interference is introduced. The B210's recall is most effected by $\alpha$ with it decreasing as $\alpha$ increases. The recall is also close to the same when the OOD interference is introduced with a loss in performance as high as 0.05.

\subsection{Discussion}
 The calibration maintains micro-recall scores close to $1-\alpha$ across gain and channel occupancy when conditioned on those dataset characteristics, with micro-recall scores as much as 0.05 below the intended score when $\alpha$ is 0.25. Additionally the calibration is robust to interference showing similar performance with the OOD radio. Across all experiments the precision increases as $\alpha$ increases. A larger $\alpha$ means allowing more false negatives with a trade-off of less false positives, with false positives having a direct relationship to precision. 

In the results for each radio it is shown that the difference in performance is due to the B210 radios whereas the models distinguishes the X310 radios well in comparison with and without the OOD interference. The X310 radios are also the closest to the receiver while the B210's are at least 20 ft away from the receiver and closer to the OOD radio. The B210 radio classes also have a sensitivity to the effects of the calibration as their performance changes as $\alpha$ changes.  

\section{Conclusions}
\label{sec:conclusions}

\label{Conclusion}
In future wireless systems, spectrum will be shared by a large number of users and devices to provide many emerging services. As a result, multiple transmissions may overlap in both time and frequency,
and identifying each transmissions (either authorized or not) becomes more challenging. 
In this study, a new framework is proposed for RF fingerprinting under co-channel interference with heterogeneous transmission protocols. Specifically, a multi-label classification problem is formulated and a lightweight multi-label RF classifier based on a 1D CNN is proposed. Furthermore, a mechanism controlling the False Negative Rate (FNR) using Conformal Risk Control for model calibration is proposed and tested. 
Results show that the model can distinguish between radios with accuracy as high as 0.97 and as low as 0.73 depending on channel conditions. Models are calibrated for the average number of False Negative to be upper bounded at 0.05, 0.15, and 0.25, with effects of the calibration shown by micro recall scores of approximately (1- calibrated false negatives) after calibration, robust to out-of-distribution interference. This work serves as a start to a larger study on RF fingerprinting under interference. Future works will include calibrated anomaly detection modeling for detecting out-of-distribution interference with statistical guarantees and considering adjacent channel interference also within the experiment.

\section{Acknowledgments}
\label{acknowledgement}
This work was supported by the US ARO under cooperative agreement W911NF-23-1-0214 and W911NF-24-2-0133, and by US NSF award 2302469, 2428761.

\bibliographystyle{IEEEtran}
\bibliography{IOTSecurityJeff}

\end{document}